# ARAACOM: ARAbic Algerian Corpus for Opinion Mining

Hichem Rahab
Laboratoire ICOSI
Université de Khenchela
(Algeria)
rahab.hichem@univ-khenchela.dz

Abdelhafid Zitouni
Laboratoire LIRE
Université de Constantine 2
(Algeria)
Abdelhafid.zitouni@univ-constantine2.dz

Mahieddine Djoudi
Laboratoire TechNE,
Université de Poitiers
France
mahieddine.djoudi@univ-poitiers.fr

## ABSTRACT
It is a challenging task to identify sentiment polarity in Arabic journals comments. Algerian daily newspapers interest more and more people in Algeria, and due to this fact they interact with it by comments they post on articles in their websites. In this paper we propose our approach to classify Arabic comments from Algerian Newspapers into positive and negative classes. Publicly-available Arabic datasets are very rare on the Web, which make it very hard to carring out studies in Arabic sentiment analysis. To reduce this gap we have created ARAACOM (ARAbic Algerian Corpus for Opinion Mining) a corpus dedicated for this work. Comments are collected from website of well-known Algerian newspaper Echorouk. For experiments two well known supervised learning classifiers Support Vector Machines (SVM) and Naïve Bayes (NB) were used, with a set of different parameters for each one. Recall, Precision and F_measure are computed for each classifier. Best results are obtained in term of precision in both SVM and NB, also the use of bigram increase the results in the two models. Compared with OCA, a well know corpus for Arabic, ARAACOM give a competitive results. Obtained results encourage us to continue with others Algerian newspaper to generalize our model.

## 1. INTRODUCTION
The proliferation of Internet use and application in our daily life we offer a large amount of data of several forms and about all domains, this treasury need a powerful means to take benefit from. A lot of available data in the web is constituted by user generated content (UGC) like product reviews, and comments submitted by users of Web sites such as Epinions.com and Amazon.com [25], also in websites of Algerian newspaper such Echorouk[1], elkhabar[2],.. etc, or television channels like Aljazeera[3] [20].

Sentiment analysis or opinion mining [15][16], is a new field in the cross road of data mining and Natural Language Processing/Natural Language Understanding (NLP/NLU)[14] which the purpose was to extract and analyze opinionated documents and classify it into positive and negative classes [4][21], or in more classes such as in [6] and [26]. Unlike data mining, where the work is to track meaningful knowledge from structured data, in sentiment analysis it is subject to find structured knowledge from unstructured amounts of data [23]. Sentiment analysis can be referenced in literature by different manners, such as: *opinion mining*, *opinion extraction*, *sentiment mining*, *subjectivity analysis*, *affect analysis*, *emotion analysis*, *review mining*, etc [15]. Application domains of sentiment analysis can include**:** product reviews, advertising systems*,* market research, public relations, financial modeling and many others [10].

A challenging task in opinion mining is comments classification to their positive and negative sentiment toward article subject. This problem will be increased in the case of Arabic language due to the morphologic complexity and the nature of comments, for instance the behavior of the reviewers could be affected by the culture in Arabic countries [21].

Arabic is one of the Semitic languages, it is written from right to left. In the Arabic language there are 28 letters from which there are three langue vowels, and eight diacritics, and there is no capitalization [12]. In today daily life, we can found three variants of Arabic language, the classical Arabic is the form used in early age of Islam religion [2], it saves their original form for centuries of years for their strongly association with Islam literature [11], that it is the language of the holy Quran [6]. The second type of Arabic language we can found is the Modern standard Arabic (MSA), which is the unified language used in all Arabic countries in education, press, and official correspondences, and it is the form we deal with in the scope of this work. While, Arabic dialects are a regional versions of Arabic spoken in daily life in Arabic countries without a written form, we can found generally tow principal types of Arabic dialects, Maghreb dialects and eastern dialects [19].

We have organized this paper as fellows: In the second section related works are presented, Third section of the paper deals with the approach we carried out for sentiment analysis, and the different steps we fellow are detailed. Experimental evaluation techniques and metrics are explained in the section four. The fifth is dedicated to present and discuss obtained results. We conclude the work in the sixth section with giving perspectives to future works.

## 2. Related Works
An important baseline to conduct studies in opinion mining is the language resources, in [21] the OCA, a publicly available corpus, is designed to implement sentiment analysis applications for the Arabic language. The authors collect 500 movie reviews from different web pages and blogs in Arabic, and they take benefit from the rating system of websites to annotate them as positive or negative, for instance, in the case of rating system of 10 points, reviews with less than 5 points are considered as negative, while those with a rating between 5 and 10 points are classified as positive. For experiments RapidMiner software is used with two machine learning algorithms, SVM and NB, who's the performances are compared, and a 10-fold cross validation method was implemented. The best results were obtained with SVM, and the use of trigram and bigram overcome the use of unigram model.

---

[1] http://www.echoroukonline.com

[2] http://www.elkhabar.com

[3] http://www.aljazeera.net

In their next work [22] the authors of OCA [21], using a Machine Translation (MT) tool, have translated the OCA corpus into English, generating the EVOCA corpus (English Version of OCA) contains the same number of positive and negative reviews. Following the work in [21] Support Vector Machines (SVM) and Naïve Bayes (NB) were applied for classification task. Obtained results are worse than OCA (90.07% of F_measure). The authors report the loss of precision in experiments with EVOCA comparing to obvious work with OCA to the quality of translation.

The authors in [4], in the goal of improving accuracy of opinion mining in Arabic language, they investigate the available OCA corpus [21], and they use the two well know machine learning algorithms SVM and NB with different parameters of SVM. Then 10, 15, and 20 fold cross validation were used. For SVM method the highest performance was obtained with Dot, Polynomial, and ANOVA kernels. Authors observe that SVM kernels: epachnenikov, mulriquadric and radial were failed at some point. For NB, its highest accuracy was achieved with BTO (Binary Term Accuracy).

In[6] and [26] a classification in five classes with SVM method was used. In [6] Arabic reviews and comments on hotels are collected from Trip Advisor website and classified into five categories: "ممتاز" (excellent); "جيد جدا" (very good); "متوسط" (middling); "ضعيف" (weak) and "مروع" (horrible), the modeling approach combined SVM with kNN provides the best result (F_measure of 97%). The authors in [26] proposed a three steps system consists of: corpus preprocessing, features extraction, and classification. The corpus used is obtained from Algerian Arabic daily Newspapers. So they focus on the second step where 20 features were used, and a combination of many SVM was used to classify comments into five classes.

The work in [3] investigate in the sentence and document levels. For the feature selection they start by the basic known feature model, the bag-of-words (BOW), where the feature model contain only the available words as attributes. A second type of feature model is created by adding the polarity score as attribute, using SentiWordNet via a machine translation step.

The work in [5] focus on Arabic tweets to study the effect of stemming and n-gram techniques to the classification process. Also the impact of feature selection on the performance of the classifier is studied. Support Vector Machines (SVM), Naïve Bayes, (NB), and K-nearest neighbor (KNN) are the used classifiers. We remark that the authors don't study the effect of changing parameters of different classifiers. In the results the authors mention that the use of feature selection technique improves significantly the accuracy of the three classifiers, and the SVM outperforms the other classifiers.

In [20] the authors used two available Arabic Corpora OCA created in [21] and ACOM which is collected from the web site of Aljazeera channel[4]. For the classification task, Naïve Bayes, Support Vector Machines and k-Nearest Neighbor were used. Stemming is investigated and the work concludes that the use of light-stemming is better than stemming. Obtained results show that the classification performance is influenced by documents length rather than the data sets size.

It is clear from this study of related works that publicly available resources for sentiment analysis in Arabic language are seldom. And those available are generally collected from movie reviews, which limit their domain of use. This fact makes it very important, for us, to create our proper corpus to be adequate with the purpose of our study.

## 3. Our Approach

In this step we will present our proposed approach for sentiment analysis in Algerian Arabic Newspaper. In our work we use RapidMiner tool kit[5], which is free for educational purpose. And we collect comments from the Algerian daily echorouk[6]. Figure 1 show the general approach we adopt.

### 3.1 Difficulties in Algerian Newspaper websites

Some difficulties are found and solved in the scope of this work:

1. A lot of spelling mistakes which is current in such texts.
2. Some comments are writing totally or partially in French language, and this remark is due to Algerian history.
3. Comments in Algerian dialect are very rare which is surprising but we can justify such a situation by the fact that comment follow the style of writing of the article.
4. The rating system give the agreement between users a strong power no matter it reflect the positive or the negative sentiment about the article.
5. The number of comments is relatively law.

### 3.2 Corpus generation

We have created our corpus for Algerian Arabic ARAACOM (ARAbic Algerian Corpus for Opinion Mining) which mean in arabic (أراؤكم, your opinions).

So we construct our corpus principally from the web site of echorouk newspaper[7]. The articles cover several topics (News, politic, sport, culture).

Compared to the important visitors of Algerian Newspaper websites, except Echorouk web site the number of comments is very low. And a lot of them are out of the main topic of the article, these comments are considered as neutral, and in the remained comments we found that negative ones largely outnumber the positives, which make for us a challenge to have equilibrium in our corpus. Table 1, present the statistics of comments collected from Echorouk web site for different categories in ARAACOM before the equilibrium between different classes. It is clear that negative ones (91) largely outnumber other categories.

---

[4] www.aljazeera.net

[5] https://rapidminer.com/

[6] www.echoroukonline.com

[7] https://www.echoroukonline.com

| Positive | Negative | Neutral | Total |
|---|---|---|---|
| 32 | 91 | 24 | **147** |

**Table 1: Number of comments from different categories in ARAACOM before equilibrium**

Another difficulty is lied to the rating system, unlike review web sites such Amazon[8] for example where user can give points in scale e.g. from 0 to 5, to the article before or after post his comment, so we can easily know for example that a user who give 5 point to article will rationally post a positive comment and one give 0 point have a negative sentiment about the article, and this very helpful in annotation process. In our case with web site Echorouk (and in the other Algerian newspaper websites) the user rather than giving points to the article, he can give one positive or negative point to other comments, without knowing who give point to which comment, this make annotation task very difficult, because having a certain number of points for a given comment have no meaning for positive or negative sentiment of the comments. So we must read carefully each comment and understand if it present a positive or negative sentiment, or even is off topic comment [17].

To generate our corpus, we take 92 comments, 60 comments from the negative category and all the 32 from the positive category (this is due to the nature of available comments -as mentioned above-) and this to have equilibrium in our corpus. The neutral comments are removed from our corpus and altered to a future work. Table 2, Table 3 and Table 4 show the statics of our corpus.

| Positive | Negative | Total |
|---|---|---|
| 32 | 60 | **92** |

**Table 2: Number of comments from different categories in ARAACOM after equilibrium**

Despite the law number of comments in our generated corpus, we continue our study, due to the fact the important number of tokens per comment is 36.45, as shown in Table 3, and is the documents length that influence in the classification performance more than the data set size [20].

| Nbre of comments | Nbre of tokens | Avg Nbre of tokens per comment |
|---|---|---|
| 92 | 3354 | 36.45 |

**Table 3: Number of tokens per comments**

|  | Positive | Negative | Total |
|---|---|---|---|
| ARAACOM | 34.78% | 65.21% | 100% |

---

[8] https://www.amazon.com

**Table 4: Statistics of percentage of docs from different categories in ARAACOM After equilibrium**

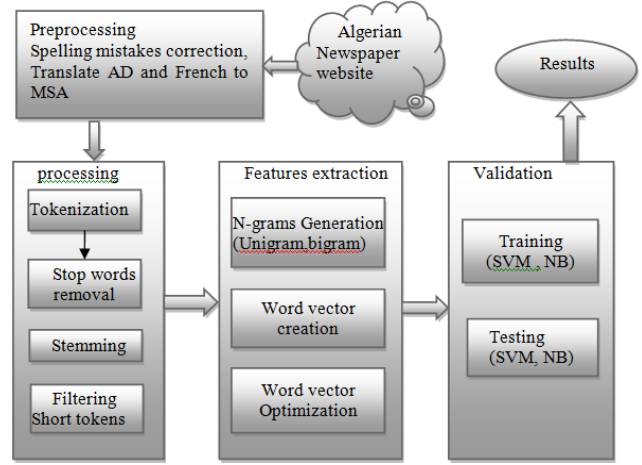

**Figure 1  Our approach general process**

## 3.3  Pre processing

To reduce the comment's vector size, some pre processing steps were conducted. Spelling mistakes were corrected manually, and words written in Algerian Dialect (AD) and in French are translated into Modern Standard Arabic (MSA), that stemming algorithms do not perform well with dialectical words and this dialectical words need an extended set of stopwords [8]. Characters encoding are resolved on UTF-8.

| Extraction from Original comment | Comment after correction |
|---|---|
| عندما يقول هذا السؤول بان البنزين غير مغشوش فلماذا اصبحنا نرانه ينفذ بسرعةهذا السوال موجه اليك لانك انت بنفسك لاتراعي لهذا لانك تعبء سيارتك باطل كل شيء من عند البايلك ولاتصرف عليه مليم واحدمن جيبك اطلب من بخابر نفطال ان تفسر لنا هذا مادمت انك تقول لاوجود للغش فبماذا تفسر ذالك ياسي ايزو | عندما يقول هذا المسؤول بان البنزين غير مغشوش فلماذا اصبحنا نرانه ينفذ بسرعة هذا السؤال موجه اليك لانك انت بنفسك لاتهتم لهذا لانك تعبيء سيارتك مجانا كل شيء من الدولة ولاتصرف عليه مليم واحد من جيبك اطلب من مخابر نفطال ان تفسر لنا هذا مادمت لاوجود للغش فبماذا تفسر ذالك ياسيد ايزو |

**Table 5: Sample of comment manual correction**

## 3.4  Comments processing

Before feature extraction, a sequence of processing steps were carried out that each comment goes through.

### 3.4.1  Tokenization

In tokenization, a stream of characters is segmented into meaningful units called tokens [3]. Each token representing a word, the splitting process can simply use spaces between words [4].

### 3.4.2  Stop Words removal

Stop words are common words to be removed from Natural Language data set before processing, due to their little impact on Opinion Mining to reduce the workspace size [4],[24]. We note that some authors such in [20] recommend that these lists should be hand crafted as it is domain and language-specific. In our work, we use Arabic stop words filter coming with RapidMiner toolkit.

### 3.4.3 Stemming
In stemming the words are reduced to their roots known as the base form or stem, without lose the most of its linguistic features [9]. There are two different stemming techniques; generally stemming simply called stemming and light-stemming. For our work we use the basic Arabic stemmer.

### 3.4.4 Filtering Short tokens
Tokens with less than two letters were removed because of their low significance in opinion mining task.

## 3.5 Feature selection
Feature selection is a process that selects a subset of original features to be used in the classification task [1]. The optimality of a feature subset is measured by some evaluation criterions [13]. We have used several feature selection parameters and different results are computed and compared.

### 3.5.1 N-grams Generation
In this work two n-grams were generated: Unigram and bigram.

### 3.5.2 Word vector creation
To create vector representing all comments, we use four different parameters:

1. **Term Frequency (TF):** A ratio representing the number of term occurrences over the total number of words.
2. **Term Frequency Inverse Document Frequency (TF-IDF):** Define the weight of a term in the context of a document [8].
3. **Term Occurrences (TO):** Each element represent the number the word occur in the comment (0 to n).
4. **Binary Term Occurrences (BTO):** The element takes 1 if the word appears at least once in the comment and 0 otherwise.

### 3.5.3 Word vector optimization
very occur words and very rare ones have a low significance in opinion mining [1]. So we eliminate words appear less than 3% in corpus, and those appear more than 30%.

## 4. Experimental Evaluations
In this section, the proposed system is evaluated. Several experiments have been accomplished. We have used cross-validation to compare the performance of two of the most widely used learning algorithms: SVM and NB.

In our experiments, the 10-fold cross-validation has been used to evaluate the classifiers.

## 4.1 Classifiers
For the classification task, two well known supervised learning methods were utilized: support vector machine (SVM) and naïve Bayes (NB).

### 4.1.1 Support vector machines
The support vector machines are a learning machine method for two-group classification problems [7]. The Support vector machines (SVM) are a widely used classifier in different disciplines, due to its ability to modeling diverse sources of data, their flexibility in handling data of high-dimensionality, and the high obtained accuracy.

### 4.1.2 Naïve Bayes
The naive bayes classifier is a well known algorithm used in text classification. In the "naive Bayes assumption" all attributes of the examples are independent of each other given the context of the class [18]. If document belongs to different classes with different probabilities, it is classified in the class that have the highest posterior probability[5].

## 4.2 Performance measures

| True class / Predictive class | Positive | Negative |
|---|---|---|
| Positive | True positive (TP) | False Positive (FP) |
| Negative | False Negative (FN) | True Negative (TN) |

**Table 6: Confusion matrix**

Table 6 show the confusion matrix where:
1. TP count the comments correctly assigned to the positive category.
2. FP count the comments incorrectly assigned to the positive category.
3. FN count the comments incorrectly rejected from the positive category.
4. TN count the comments correctly rejected from the positive category.

Performance measures are defined and computed from this table; three parameters were used, precision, recall, and $F_1$_measure (or too simply F_mesure).

$$Precision = \frac{TP}{TP + FP}$$

$$Recall = \frac{TP}{TP + FN}$$

Precision and recall are complementary one to the other, we combine the two using the $F_1$ measure called generally $F_1$, given as:

$$F_1\_measure = 2 \times \frac{Precision \times Recall}{Precision + Recall}$$

## 5. Results and discussion
### 5.1 Evaluation with Support Vector Machines
In the scope of our work, we have applied the SVM algorithm with two different kernel types, Anova and polynomial. And both Unigram and Bigram models were tested. For the comments vector creation four variants was implemented, Term occurrence (TO), Term Frequency (TF), Term Frequency Inverse Document Frequency (TF-IDF), and Binary Term Frequency (BTO).

Best results are obtained in precision when TF and TF-IDF are suited for vector creation and this for both SVM implemented kernels. The use of bigram increase results in most of cases.

|  | Kernel | F_measure | precision | Recall |
|---|---|---|---|---|
| Term Occurrence | Anova | 79,64% | 88,33% | 72,50% |
|  | Polynomial | 61,19% | 47,80% | 85,00% |
| Term Frequency | Anova | 82,86% | **96,67%** | 72,50% |
|  | Polynomial | 67,16% | 88,33% | 54,17% |
| TF-IDF | Anova | 81,78% | **100%** | 69,17% |
|  | Polynomial | 70,11% | **96,67%** | 55,00% |
| Binary Term Occurrence | Anova | 83,00% | 91,67% | 75,83% |
|  | Polynomial | 69,58% | 57,40% | 88,33% |

**Table 7: Results with SVM using Uni-gram model**

|  | Kernel | F_measure | precision | Recall |
|---|---|---|---|---|
| Term Occurrence | Anova | 79,45% | 93,33% | 69,17% |
|  | Polynomial | 62,93% | 48,87% | 88,33% |
| Term Frequency | Anova | 83,50% | **100%** | 71,67% |
|  | Polynomial | 75,00% | **100%** | 60,00% |
| TF-IDF | Anova | 79,39% | **100%** | 65,83% |
|  | Polynomial | 63,64% | **100%** | 46,67% |
| Binary Term Occurrence | Anova | 84,78% | **97,50%** | 75,00% |
|  | Polynomial | 68,31% | 55,69% | 88,33% |

**Table 8: Results with SVM using Bi-gram model**

For comparing the results of our ARAACOM corpus with OCA used in [21], we choose the term frequency vector and the anova kernel for SVM, and this because the work with OCA in[21] does not testing several SVM parameters.

|  | N-gram Model | Precision | Recall | Other Metrics |
|---|---|---|---|---|
| ARAACOM | Unigram | **96,67%** | 72,50% | F_measure =82,86% |
|  | Bigram | **100%** | 71,67% | F_measure =83,50% |
| OCA in [21] | Unigram | 86.99% | 95.20% | ACC= 90.20% |
|  | Bigram | 87.38% | 95.20% | ACC= 90.60% |

**Table 9: Comparison between ARAACOM And OCA in SVM Classification**

The results show that in term of precision ARAACOM outperform OCA, which mean that our predictive results are more important. In the other hand the recall of OCA is better than ARAACOM which indicate that OCA documents are well classified than ARAACOM ones.

### 5.2 Evaluation with Naïve Bayes
As with SVM best results are found in precision, with the different vector creation methods. And the use of bigram also increases the obtained results.

|  | F_measure | precision | recall |
|---|---|---|---|
| Term Occurrence | 80,05% | **95,00%** | 69,17% |
| Term Frequency | 80,05% | **95,00%** | 69,17% |
| TF-IDF | 72,15% | 90,48% | 60,00% |
| Binary Term Occurrence | 80,93% | **97,50%** | 69,17% |

**Table 10: Results with Naïve Bayes using Uni-gram model**

|  | F_measure | precision | recall |
|---|---|---|---|
| Term Occurrence | 80,93% | **97,50%** | 69,17% |
| Term Frequency | 80,93% | **97,50%** | 69,17% |
| TF-IDF | 73,55% | **95,00%** | 60,00% |
| Binary Term Occurrence | 81,78% | **100%** | 69,17% |

**Table 11: Results with Naïve Bayes using Bi-gram model**

As for the SVM, we will compare our the results of our corpus in classification with Naïve Bayes classifier with OCA corpus [21].

|  | n-gram Model | Precision | Recall | Other Metrics |
|---|---|---|---|---|
| ARAACOM | Unigram | **97,50%** | 69,17% | F_measure = 80,93% |
|  | Bigram | 81,78% | **100%** | F_measure = 69,17% |
| OCA in [21] | Unigram | 79.99% | 85.60% | Acc = 81.80% |
|  | Bigram | 82.75% | 88.80% | Acc = 84.60% |

**Table 12: Comparison between ARAACOM And OCA in NB Classification**

In this case our system ARAACOM outperforms OCA in term of precision when using Uigram model, and in term of recall when using the bigram model.

## 6. Conclusion and future works

Our exploration of obvious achieved works in sentiment analysis, especially in Arabic language, show the lack in publicly available resources dedicated for carried out Arabic sentiment analysis studies. And in these rarely available ones, we found that the most are interested by movie reviews; perhaps because of the important number of web sites inciting the public to review films and serials. Such available resources are not adapted to use in other domains such as newspaper comments sentiment analysis which cover several topics like politics, culture, sports, medicine, etc.

In this work we present our approach of sentiment analysis in Algerian Arabic daily newspapers. The approach starts with the corpus creation where 92 comments are collected from echorouk newspaper web site. And ARAACOM (ARAbic Algerian Corpus for Opinion Mining) was created to be used in the scope of this study. Some processing operations were conducted.

Despite difficulties found in Algerian daily newspaper website, as the language mistakes, the negative sentiment of the majority of comments and also the rating system that does not at all help us in the annotation step, we overcome these obstacles to create a some equilibrate corpus ARAACOM.

Comments are represented in different vector models, TO, TF, BTO and TF IDF, also with unigram and bigram models. For classification, two well known methods are used, support vector machines SVM and naïve bayes NB. And different parameters for each classifier were tested. In the validation process 10-fold cross validation method was conducted to train and test the models.

Obtained results are very promising, in term of precision and recall. But in term of F_measure as a compromise between precision and recall the results remain modest which need more work to improve this rate.

Compared with the well know corpus for Arabic OCA, our approach ARAACOM give a competitive results, for the SVM model ARAACOM outperform OCA in classification precision. These results encourage us to continue in this issue in future works.

As perspective to this work we would to add the neutral class which allow us using all available comments (or at least an important part of them), so other methods can be implemented or a combination of existing methods can be used to resolve the multi classes problem. Also the corpus must be enriched by comments from other Algerian newspaper to generalize the model.

Another point is to dealing with comments written in Algerian dialect and French language directly without the need of the manual translation. In this point machine translation may be envisaged as an automated process.

Also it is very important to take in consideration in future works, the article topic when searching the sentiment orientation of comments which can allow us considering more comments in the classification step, that a lot of comments are removed from our corpus due to the fact that their semantic orientation is strongly related to the article topic.